\documentclass[11pt,a4paper]{article}
\usepackage[hyperref]{acl2021}
\usepackage{times}
\usepackage{latexsym}

\usepackage{microtype}

\usepackage{amsmath}
\usepackage{pgf-pie}
\usepackage{algorithm}
\usepackage[noend]{algpseudocode}
\usepackage{caption}
\usepackage{subcaption}
\usepackage{amsfonts}
\usepackage{tablefootnote}
\usepackage{pifont}
\newcommand{\cmark}{\ding{51}}%
\newcommand{\xmark}{\ding{55}}%

\aclfinalcopy 

\makeatletter
\newcommand{\printfnsymbol}[1]{%
  \textsuperscript{\@fnsymbol{#1}}%
}
\patchcmd{\@verbatim}
  {\verbatim@font}
  {\verbatim@font\small}
  {}{}
\makeatother

\usepackage{varwidth}

\title{{F}ast{S}eq: Make Sequence Generation Faster}

\author{Yu Yan\textsuperscript{1} \thanks{\:\:\:Equal contribution}\:\:, Fei Hu\textsuperscript{1} \printfnsymbol{1}, Jiusheng Chen\textsuperscript{1} \printfnsymbol{1}\thanks{\:\:\:Corresponding author}\:\:, Nikhil Bhendawade\textsuperscript{1} \printfnsymbol{1}, Ting Ye\textsuperscript{1}, \\
\textbf{Yeyun Gong\textsuperscript{2}, Nan Duan\textsuperscript{2}, Desheng Cui\textsuperscript{1}, Bingyu Chi\textsuperscript{1} \and Ruofei Zhang\textsuperscript{1}} \\
  \textsuperscript{1}Microsoft, \textsuperscript{2} Microsoft Research Asia \\
  $\left\{\begin{varwidth}{12cm}\centering \texttt{yyua,fhu,jiuchen,nibhenda,tiy,}\\ 
  \texttt{yegong,nanduan,decu,bingychi,bzhang}\end{varwidth}\right\}$\texttt{@microsoft.com}
  }

\date{}

\begin{document}
\maketitle
\begin{abstract}
Transformer-based models have made tremendous impacts in natural language generation. However the inference speed is a bottleneck due to large model size and intensive computing involved in auto-regressive decoding process. We develop FastSeq framework to accelerate sequence generation without accuracy loss. The proposed optimization techniques include an attention cache optimization, an efficient algorithm for detecting repeated n-grams, and an asynchronous generation pipeline with parallel I/O. These optimizations are general enough to be applicable to  Transformer-based models (e.g., T5, GPT2, and UniLM). Our benchmark results on a set of widely used and diverse models demonstrate 4-9x inference speed gain. Additionally, FastSeq is easy to use with a simple one-line code change. The source code is available at \url{https://github.com/microsoft/fastseq}.
\end{abstract}

\section{Introduction}
Transformer-based model architectures have made tremendous impact in multiple domains. However, due to large model size and intensive computing involved in the decoding process, the inference speed is still a bottleneck for long sequences applications \citep{wu2016googles, tay2020efficient}. 
A variety of model architectural innovations have been proposed to increase the generation speed from different perspectives. One trend is to change the model architectures, like model distillation~\citep{shleifer2020pre} and sparse attention~\citep{beltagy2020longformer}. Although these techniques can alleviate the performance issue, there may be still some trade-off between model accuracy and speed. On the other hand, efficient infrastructures have been developed to accelerate the inference speed, e.g., TensorRT ~\citep{vanholder2016efficient} and FasterTransformers\footnote{\href{https://github.com/NVIDIA/DeepLearningExamples/tree/master/FasterTransformer}{FasterTransformer Github}}.

In this paper, we present FastSeq framework to make sequence generation faster. FastSeq can accelerate the sequence generation by 4x to 9x with a simple one-line code change for models in FairSeq~\citep{ott-etal-2019-fairseq} and Huggingface-Transformers~\citep{wolf-etal-2020-transformers}. The design principle of FastSeq is to improve the inference speed without losing model accuracy and usability.

Our optimization approaches include an attention cache optimization, an efficient algorithm for detecting repeated n-grams, and an asynchronous generation pipeline with parallel I/O. These optimizations are general enough for a wide range of Transformer-based model~\citep{NIPS2017_3f5ee243} architectures, including the encoder-decoder architecture (e.g., T5~\citealt{JMLR:v21:20-074}, BART~\citealt{lewis-etal-2020-bart}, ProphetNet~\citealt{qi-etal-2020-prophetnet}), the decoder-only architecture (e.g., GPT2~\citealt{radford2019language}), and the encoder-only architecture (e.g., UniLM~\citealt{NEURIPS2019_c20bb2d9}).
FastSeq is also designed to be flexible for extension on supporting other models and frameworks. 
Our technologies are partially adopted by FairSeq\footnote{See pull requests \href{https://github.com/pytorch/fairseq/commit/bff7f85206f6f64b9455035893d44d66b98e33b0}{FastSeq n-gram Blocking} and \href{https://github.com/pytorch/fairseq/pull/1852}{Beam Search Perf Improvement}}.
A demo video can be found at  \url{https://www.youtube.com/watch?v=jrdsEUxhSEE}.

\section{Preliminary Analysis} \label{pre_analysis}
For models with similar size, the sequence generation is much slower than classification, regression or language score computation. 
Why is the generation so time-consuming? Before analyzing the reasons, let's recap the generation algorithms first.

\subsection{Generation Algorithms}
Encoder-decoder structure is used in the most competitive models for sequence-to-sequence generation. The encoder side takes an input sequence of symbol representations $(x_{1},..., x_{n})$ and outputs a sequence of continuous representations $ \textbf{z} = (z_{1},..., z_{n})$. Then the decoder side generates an output sequence $(y_{1},..., y_{t})$ with one element at a time. At each step, the model is auto-regressive by consuming the previously generated symbols and then computing the probability scores to select the next element. Greedy search and beam search are two popular algorithms used for the selection of next element. The difference between them is that at each step, greedy search only selects one candidate with maximum score, but beam search selects the top $k$ candidates as beams. As beam search maintains multiple beams during the generation, it usually outputs a better result than greedy search.

To avoid repeated computation in the attention layer, the key ($K$) and value ($V$) from previous and current steps are usually cached to compute the next token. Equation~\eqref{eq.attn} describes how the self-attention with the cache mechanism is implemented at step $t$. 

\begin{equation}
\begin{aligned}
\underset{\scriptscriptstyle{[B \times M, 1, D]}}{Q_{t}} = \underset{\scriptscriptstyle{[B \times M, 1, D]}}{y_{t-1}} \cdot \underset{\scriptscriptstyle{[D \times D]}}{W_{q}} \\
\underset{\scriptscriptstyle{[B \times M, t, D]}}{K_{t}} =  concat(\underset{\scriptscriptstyle{[B \times M, t-1, D]}}{Cache\_K_{t-1}}, {y_{t-1}} \cdot \underset{\scriptscriptstyle{[D \times D]}}{W_{k}}) \\
\underset{\scriptscriptstyle{[B \times M, t, D]}}{V_{t}} =  concat(\underset{\scriptscriptstyle{[B \times M, t-1, D]}}{Cache\_V_{t-1}}, {y_{t-1}} \cdot \underset{\scriptscriptstyle{[D \times D]}}{W_{v}}) \\
\underset{\scriptscriptstyle{[B \times M, 1, D]}}{attn_t} = softmax(\dfrac{{Q_{t}}{K_{t}^T}}{\sqrt{d_{k_{t}}}}){V_{t}} \\
\end{aligned}
\label{eq.attn} 
\end{equation}

where $B$ is the batch size; $M$ is the beam size; $D$ is the embedding dimension; $Q_{t}$, $K_{t}$, $V_{t}$ represent \textbf{query}, \textbf{key}, \textbf{value} respectively, and are in the shape of $\mathbb{R}^{B \times M} \times \mathbb{R}^T \times \mathbb{R}^D$; $W_{q}$, $W_{k}$, $W_{v}$ are the weights for the query, key, and value in the shape of $\mathbb{R}^{D\times D}$; $attn_t$ is in the shape of $\mathbb{R}^{B \times M} \times \mathbb{R}^1 \times \mathbb{R}^{D}$.

To simplify the equations, we do not consider multi-heads here, but these equations can be adjusted to be of multi-head style.

\subsection{Bottlenecks in Generation}\label{bottlenecks}
\begin{figure}[H]
\begin{subfigure}[t]{3in}
    \centering
	\includegraphics[width=2.0in]{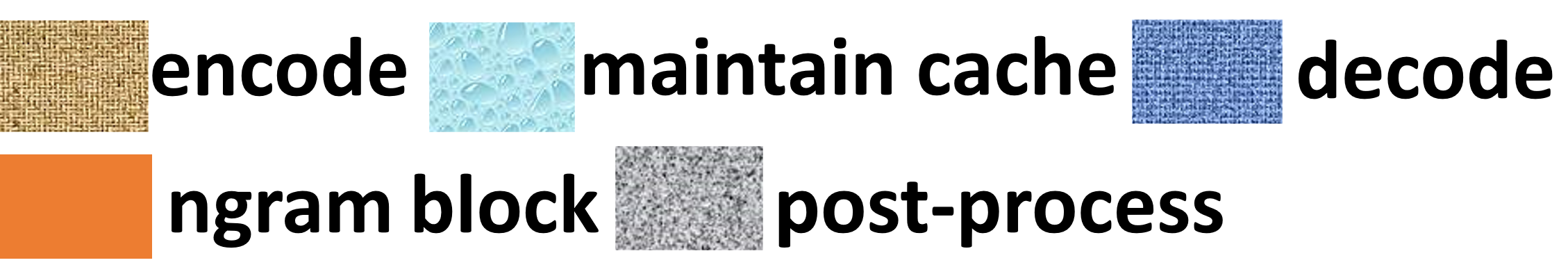}
\end{subfigure}
\begin{subfigure}[t]{1.5in}
    \centering
	\includegraphics[width=1.35in]{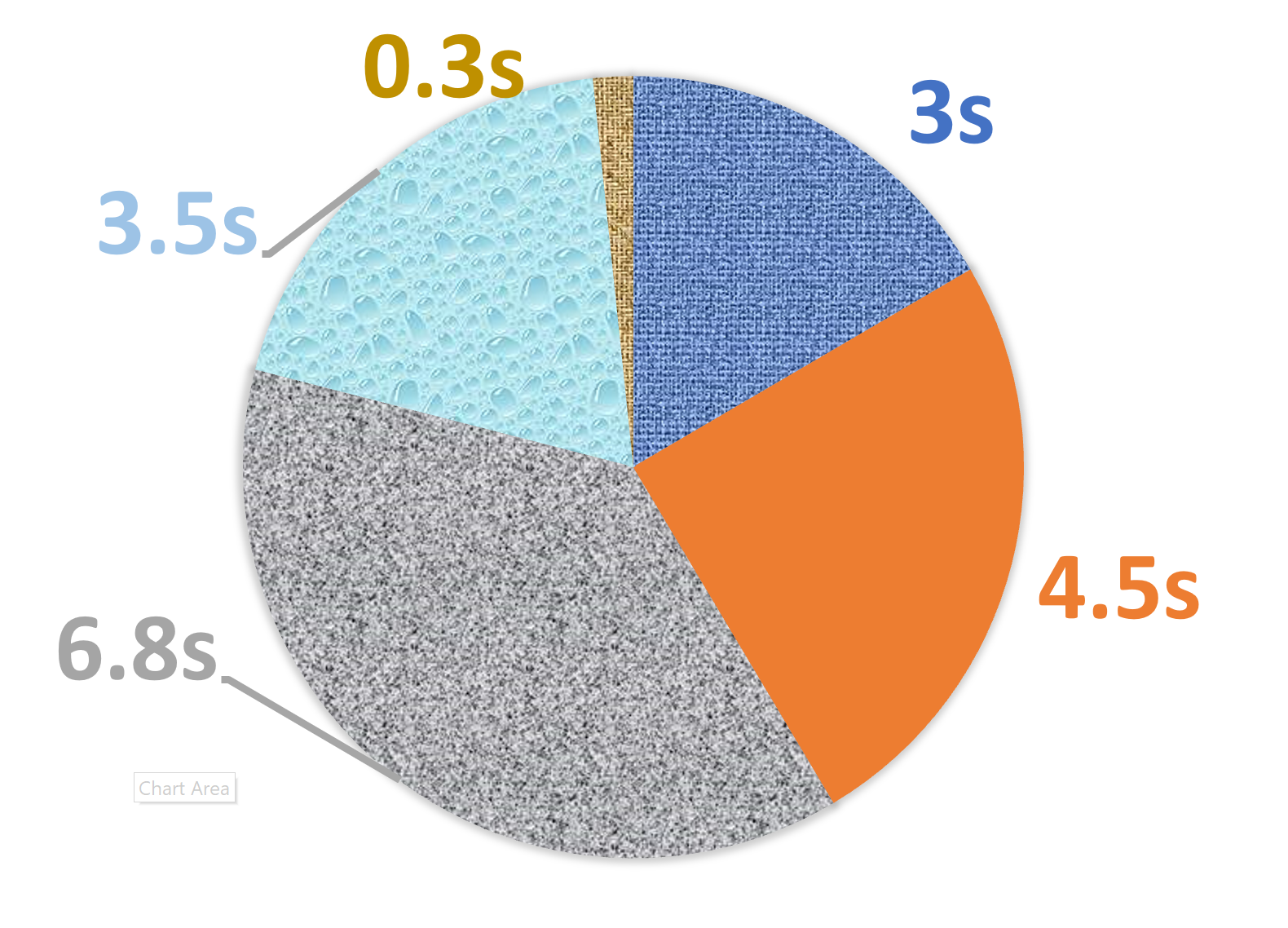}
\caption{Before optimizations}\label{figure.before_opt}
\end{subfigure}
\begin{subfigure}[t]{1.15in}
    \centering
    \raisebox{0.045in}{
        \includegraphics[width=1.45in]{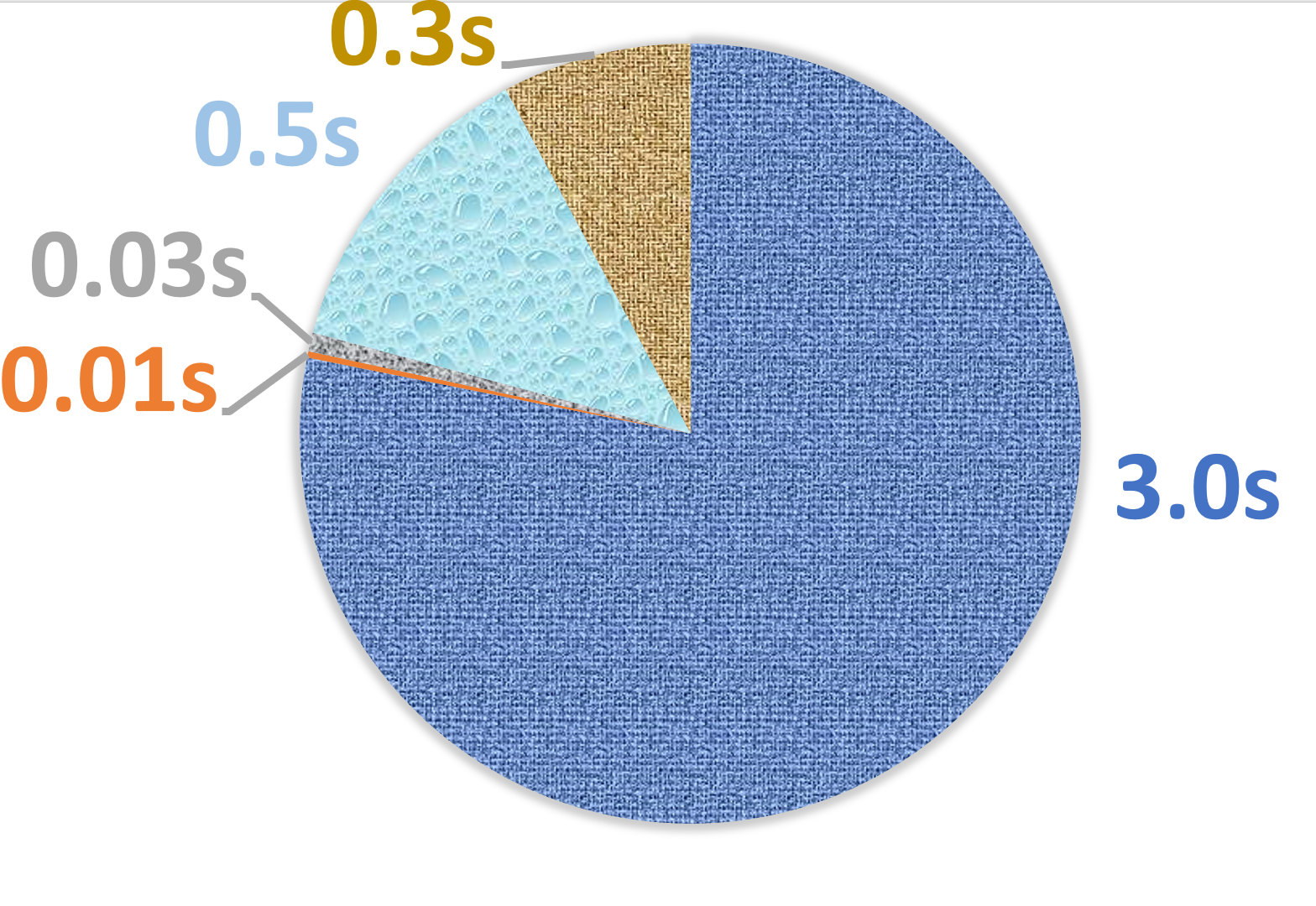}
    }
\caption{After optimizations}\label{figure.after_opt}
\end{subfigure}
\caption{(a) Before optimizations: non-computation operations, e.g, maintain cache, n-gram blocking and post-process cost most of the time. (b) After optimizations: majority of time is spent on encode and decode. }
\label{figure.time_break}
\end{figure}

Figure~\ref{figure.before_opt} shows the profiling results of running the official BART model implemented by FairSeq. It indicates that maintaining cache, blocking n-gram repeats, and post-process individually take longer time than decoding itself. Profiling is done by running the official BART implemented by FairSeq v0.0.9 on CNN DM dataset with default parameters (batch size 32, beam size 4, and no-repeat n-gram 3). Non-computation parts, like maintain cache,  blocking n-gram repeats and post-process, cost more than 80\% of the generation time. 
We analyze these time-consuming components below.

\paragraph{Cache Maintenance} Along with better generation results, beam search introduces significant additional computational and memory cost. As Equation~\eqref{eq.attn} indicates, the size of $X_t$, $Q_t$, $K_t$, $V_t$, and $attn_t$ in beam search is $M$ times larger than those in greedy search. It results in more memory consumption, larger matrix operations (e.g., $concat$), and more expensive cache maintenance (e.g., reordering the top-$k$ beams and the cached key and value at each step). Moreover, the batch size is constrained by large occupied memory, which results in a low GPU utilization.

\paragraph{Block N-Gram Repeats} Blocking N-Gram Repeats is a widely used operation to avoid an n-gram appears more than once in natural language model \citep{paulus2018a, klein-etal-2017-opennmt}. It prohibits the repetitive generation of n-grams by setting their probability scores to zero. However, conventional implementation often needs to scan text sequentially and move data between GPU and CPU frequently. Its time complexity is quadratic in terms of sequence length. When processing long sequences, this operation becomes another bottleneck.

\paragraph{Post-process} It deals with detokenization and final result output. Post-process performance is largely restricted by two parts: frequent exchange of small data between GPU and CPU and the detokenization efficiency. 
In addition, for a synchronized pipeline, post-process will block the generation for the next batch of samples, while there is no required dependency between these two components. 

\section{Design}
In order to address above bottlenecks, optimizations need to be done at multiple levels, including operations, models, and pipelines, which basically touch every component of a sequence generation framework. It is a non-trivial burden for researchers and practitioners. As a result, we develop this FastSeq library to address these barriers and speed up end-to-end inference in sequence generation. FastSeq is designed with following features: (i) speed up the inference of sequence models without any accuracy loss; (ii) easy to use and compatible Python APIs with FairSeq and HuggingFace-Transformers; (iii) flexible to be extended to support new models and frameworks.  

FastSeq is written in PyTorch~\citep{NEURIPS2019_bdbca288} and composed of (1) \textbf{ops} module: provide efficient implementations of kernels (e.g., block n-gram repeats); (2) \textbf{optimizer} module: optimize model implementations in run-time, where more efficient implementations will be automatically patched to replace the ones in existing NLP toolkits (e.g., FairSeq and HuggingFace-Transformers) or the deep learning libraries (e.g., PyTorch); (3) \textbf{models} module: define the model architectures (e.g., ProphetNet, UniLM). It is noteworthy that the models in FairSeq and HuggingFace-Transformers are natively supported as well. Only one-line code change is needed to make them work with FastSeq; (4) \textbf{command line interfaces (CLIs)} module: run the inference via commands with an asynchronous pipeline, including preprocess (e.g., tokenization), generation process, and post-process (e.g., detokenization). These CLIs are compatible with FairSeq and HuggingFace-Transformers as well. Users can use the same parameters to run their end-to-end inferences. 

FastSeq is designed to be easy to use. Existing model usages (e.g., model content and parameter settings) in FairSeq and Huggingface-Transformers do not need to be changed. The example code can be found in below:
\begin{itemize}
\item Python API
\end{itemize}
\begin{verbatim}
# simply add the import of FastSeq

import fastseq
import torch

bart = torch.hub.load(
    'pytorch/fairseq',
    'bart.large.cnn')

bart.cuda().eval().half()
slines = [
    "Welcome to FastSeq. "
    "Hope you enjoy it."]
hypotheses = bart.sample(
    slines, 
    beam=4, 
    lenpen=2.0, 
    max_len_b=140, 
    min_len=55, 
    no_repeat_ngram_size=3)
print(hypotheses)
\end{verbatim}

\begin{itemize}
\item  Command Line Interface
\end{itemize}

\begin{verbatim}
fastseq-generate-for-fairseq \
      DATA \
      --path MODEL \
      --fp16 \
      --task translation \
      --batch-size BATCH_SIZE \
      --gen-subset valid \
      --bpe gpt2 \
      --beam 4 \
      ...
\end{verbatim}

\section{Optimizations}
\label{optimizations}
To address the bottlenecks discovered in Section~\ref{bottlenecks}, we develop following optimizations.

\subsection{Attention Cache Optimization} \label{attn_cache_opt}
This section introduces how the cache for the key and value in self-attention and encoder-decoder attention can be optimized to further speed up the inference.
We describe the cache deduplication below, see more comprehensive analysis and a new attention method with faster speed in our work EL-Attention~\citep{yan2021attention}

\subsubsection{Cache Optimization in Self-Attention} \label{self_attn_cache_opt}
For the decoder-only or encoder-only Transformer models (e.g., GPT2, UniLM), $X$ is the prefix of the generated hypothesis. In conventional implementations, $X$ is replicated along beam dimension, and the corresponding partial in the key ($K$) and value ($V$) is same for each beam. This means, assuming $K_t$ and $V_t$ to be of dimension $[B, M, N + T, D]$, ${K_0}(b, i, n, d) = \cdots = {K_t}(b, j, n, d)$ and ${V_0}(b, i, n, d) = \cdots = {V_t}(b, j, n, d)$, for $\forall b \in [0, B)$, $\forall i,j \in [0, M)$, $\forall n \in [0, N)$, $\forall d \in [0, D)$, where $N$ is the length of $X$, $B$ is the batch size, $M$ is the beam size, $D$ is the embedding dimension.

To optimize the cache in self-attention, we can split the cached key and value in Equation~\eqref{eq.attn} in two parts: $Cache\_K'$ and $Cache\_V'$ for the prefix; $Cache\_K_{t}$ and $Cache\_V_{t}$ for the generated sequence up till the time step $t$. With this split, the size of $Cache\_K'$ and $Cache\_V'$ can be reduced from $B \times M \times N \times D$ to $B \times 1 \times N \times D$. This also helps decrease cache reorder complexity by a factor of $M$. 

However, the above split operation results in incompatible shapes between $Cache\_K'$ and $Cache\_K_{t}$, and between $Cache\_V'$ and $Cache\_V_{t}$. Instead of reshaping these cached keys and values, $einsum$ is utilized to compute $attn_t$. This way, the expensive $concat$ operations on large tensors can be avoided.

With the above changes, the matrix operations will be conducted on the tensors with much smaller size, so the peak memory can be smaller, the operations can run faster, and then a larger batch size can be leveraged. For example, at the step $t$, the sizes of $Cache\_K_{t-1}$ and $Cache\_V_{t-1}$ decrease from $B \times M \times (N + t - 1) \times D$ to $B \times M \times (t - 1) \times D$ by $\frac{N + t - 1}{t - 1}$ times. Then $concat(Cache\_K_{t-1}, x_t \cdot W_{k})$ and $concat(Cache\_V_{t-1}, x_t \cdot W_{V})$ can be much quicker than before due to less GPU memory allocation, copy, and deallocation. The peak memory during $concat$ is largely reduced as well. Meanwhile, this implementation will save the same amount of data movement when reordering the beams in $Cache\_K_{t-1}$ and $Cache\_V_{t-1}$ because $Cache\_K'$ and $Cache\_V'$ do not need to be frequently reordered since they are de-duplicated along beam dimension.

\subsubsection{Cache Optimization in Encoder-Decoder Attention}
The cached key and value in the encoder-decoder attention also have duplication. The reason is that the key and value in the encoder-decoder attention are calculated based on the final output hidden state ($S$) from the encoder side. Accordingly, the elements of cached key and value at the beam dimension are the same. Therefore, the size of $Cache\_K$ and $Cache\_V$ can be reduced by $M$ times, from $B \times M \times N \times D$ to $B \times 1 \times N \times D$. Then the optimization benefits mentioned in Section~\ref{self_attn_cache_opt} can be achieved here as well, including peak memory reduction and larger batch size. Additionally, the cached key and value are not needed to be frequently reordered since the elements at the beam dimension are exactly the same.

Notably, the above proposed optimizations are general and can be applied to a variety of models with different architectures if they share following features: 1) attention-based architectures, including self-attention or encoder-decoder attention; 2) auto-regressive decoding based on beam search. These models could be classic Transformer-based encoder-decoder architectures (e.g., BART, ProphetNet, T5), Transformer-based decoder-only architectures (e.g, GPT2), or Transformer-based encoder-only architectures (e.g., UniLM).

The detailed implementations of the optimized self-attention and encoder-decoder attention is provided in the Appendix.

\begin{table*}
\centering
\begin{tabular}{lllccc}
\hline 
Model & Architecture & Task & Baseline & FastSeq & Speedup \\ \hline
\multicolumn{6}{c}{\textit{encoder-decoder architecture}} \\ \hline
BART~\citep{lewis-etal-2020-bart} & 12L-12L-1024 & CNN/DailyMail & 2.4 & 18.4 & 7.7x \\
DistilBART (\citeauthor{wolf-etal-2020-transformers}) & 12L-6L-1024 & CNN/DailyMail & 3.4 & 18.5 & 5.4x \\
ProphetNet~\citep{qi-etal-2020-prophetnet} & 12L-12L-1024 & CNN/DailyMail & 2.8 & 10.7 & 3.8x \\
T5~\citep{JMLR:v21:20-074} & 12L-12L-768 & WMT16 EN-RO & 8.7 & 31.3 & 4.3x \\
Transformer~\citep{ott-etal-2018-scaling} & 6L-6L-1024 & WMT16 EN-DE & 96.0 & 417.0 & 4.3x \\
\hline
\multicolumn{6}{c}{\textit{decoder-only architecture}} \\ \hline
GPT2~\citep{radford2019language} & 0L-12L-768 & CNN/DailyMail & 3.0 & 16.7 & 5.5x \\
\hline
\multicolumn{6}{c}{\textit{encoder-only architecture}} \\ \hline
UniLM~\citep{NEURIPS2019_c20bb2d9} & 12L-0L-768 & CNN/DailyMail & 1.7 & 16.4 & 9.6x \\
\hline
\end{tabular}
\caption{\label{table.e2e} Benchmark results on models of different architectures. Speed is measured by samples/s. } 
\end{table*}

\subsection{GPU-based Block N-Gram Repeats Algorithm} \label{sec_n-gram_block}
\begin{algorithm}[tb]
 \caption{GPU version no-repeat-ngram algorithm with arguments -  ngram length $n$, previously generated tokens $tokens$, current step token probability distribution $probs$.}
 \label{algo.ngram}
 \begin{algorithmic}
    \Function{block}{$tokens, probs, n$}
    \State $nBlk = tokens.rows$
    \State $nThr = tokens.columns + 1 - n$
    \State $shMem = sizeof(tokens.row(0))$
    \State $\textrm{BAN}<<<nBlk , nThr , shMem >>>$ 
    \State $(tokens, probs, n)$
    \EndFunction

    \State \:

    \Function{ban}{$tokens, probs, n$}
    \State $row = blockIdx.x$
    \State copy $row$-th row of $tokens$ from global 
    \State mem to shared mem $shm$
    \State $col = threadIdx.x$
    \State $start = tokens.columns + 1 - n$
    \For{$i=0$ {\bfseries to} $n-1$}
       \If{$shm[col+i] \neq shm[start+i]$}
       \State return
       \EndIf
    \EndFor

    \State $tokenToBan = shm[col + n -1]$
    \State $probs[row, tokenToBan] = 0$
    \EndFunction
\end{algorithmic}
\end{algorithm}
As observed in Figure~\ref{figure.before_opt}, the cost of block n-gram repeats algorithm is as high as 25\% of generation time. To reduce the cost, a new GPU-based kernel (see Algorithm~\ref{algo.ngram}) is developed to leverage the power of parallel compute and achieves the following benefits: 
1) avoiding data movement between GPU and CPU 
to alleviate the throughput bottleneck of PCIe bus interface. 
2) scanning n-grams in parallel. Instead of sequentially scanning tokens for detecting repeated n-grams, they can be scanned in parallel using threads equal to the number of n-grams generated till the time step $t$. Furthermore, each sample in a batch can be processed in parallel using multiple thread-blocks.
3) using GPU shared memory for faster memory access.

Since each token needs to be read multiple times (equal to token length of n-gram), they are stored in shared memory instead of global memory for faster access.  
\citet{jia2018dissecting} reports shared memory bandwidth for Volta V100 is 16x of global memory bandwidth.
Although there are multiple ways to organize CUDA thread blocks, our approach is to assign each n-gram to a thread and each thread-block to handle a sequence stream. In this way, Block N-gram repeats is parallelized along horizontal and vertical dimensions of a batch.

\subsection{Asynchronous Pipeline with Parallel I/O}
As shown in Figure \ref{figure.before_opt}, post-process takes significant time (6.8s) in the generation process. It is under-optimized in many existing seq2seq frameworks. One reason is that post-process is not a part of the training process, many efforts are spent on optimizing the training pipeline and the model structure rather than the generation speed. 
Another reason is, despite of works focusing on generation speed, like distilling model, the speed metric only covers the computation time but does not include the post-process part. 
For example, FairSeq does not consider the post-process time when it measures the speed. These biases result in a big overlooked speed-up opportunity. 

To improve the efficiency of the pipeline, we develop an asynchronous pipeline with parallel I/O. 
Similar to pre-fetch technology which loads next batch of data to GPU while running inference on the current batch, we post-process the current batch in a background thread while running generation on the next batch.

\section{Evaluation}
In the benchmarks, FairSeq and HuggingFace-Transformers are used as the baseline to evaluate the performance. The selected models cover different kinds of architectures, including the encoder-decoder models (e.g., BART, DistilBART, T5, ProphetNet), the decoder-only models (e.g., GPT2), and the encoder-only models (e.g., UniLM). CNN / Daily Mail dataset~\citep{NIPS2015_afdec700} and WMT'16~\citep{bojar-etal-2016-findings} are used as the benchmark datasets. The benchmark experiments are split into two groups 1) HuggingFace-Transformers with/without FastSeq; 2) FairSeq with/without FastSeq. If both FairSeq and HuggingFace-Transformers have implemented the model, we choose the faster result as the baseline. 

\paragraph{Hardware} The experiments are conducted on a node with 1 GPU (NVIDIA Tesla V100 PCIe 16GB ) and 24 cores CPU (Intel(R) Xeon(R) CPU E5-2690 v4 @ 2.60GHz).

\subsection{End-to-end Performance} 
The end-to-end benchmarks (including model loading, preprocess, model inference, and post-process) have been conducted to evaluate the performance. For each model, we use the same configuration except batch size. We search the largest batch size for each framework by doubling it per search run. Each experiment is executed 10 times and the average running time is computed as the final result. The speed number is measured in samples per second.

With the optimizations of FastSeq, the end-to-end performance yields a roughly 4x to 9x speedup, see Table~\ref{table.e2e} for more details\footnote{The baseline for BART is FairSeq and the baseline for DistilBART is Huggingface Transformers.}. In the baseline, for summarization dataset CNN/DailyMail, the speed of all models (e.g., BART, DistilBART, ProphetNet, GPT2, UniLM) is between 1.7 and 3.4 samples per second. 
Enabling FastSeq boosts the speed to more than 10 samples per second for all models studied here, and the BART model achieves 18.4 samples per second, which is 7.7 times speedup. 
On the two WMT16 translation datasets, FastSeq improves throughput by 4.3 times.

In following sections, we will present analyses on the three optimizations used in FastSeq.

\subsection{Analysis of the Cache Optimization}
To evaluate effect of the cache optimizations introduced in Section~\ref{attn_cache_opt}, Table~\ref{table.perf.analysis} compares the results of not using cache, using conventional cache, and using the proposed optimized cache. Although the computing complexity is the same for both cache-based approaches, the proposed cache optimization approach reduces the usage of GPU memory by 3.5 times. Such smaller cache memory can speed up $concat$ operations and reduce the data movement during the beam reordering, and also allow a larger batch size. These advantages together increase generation throughput from 5.6 to 18.4 samples/s.

\begin{table}
\small
\centering
\begin{tabular}{l|c|cc}
\hline
Model & Batch  & Cache & Throughput    \\
 & size & GB & samples/s  \\
\hline
BART\textsubscript{large} no cache & 32  & 0.0 & 1.8 (0.7x) \\
BART\textsubscript{large} & 32  & 6.3 & 2.4 (1.0x) \\
\hline
+Asynchronous pipeline & 32  & 6.3 & 3.6 (1.5x) \\
+GPU n-gram block & 32  & 6.3 & 5.6 (2.3x) \\
+Attention cache optimize & 32  & 1.8 & 8.1 (3.3x) \\
+Larger batch & 128  & 7.2 & 18.4 (7.7x) \\
\hline
\end{tabular}
\caption{\label{table.perf.analysis} BART\textsubscript{large} is the official version from FairSeq. No cache: disable cache on FairSeq. Generation parameters: beam size = 4, no-repeat n-gram = 3. Data: CNN DM validation dataset. Cache size is estimated according to max input length 1024, output length 50.} 
\end{table}

\subsection{Analysis of Block N-Gram Repeats}
To demonstrate the effectiveness of GPU kernel described in Section~\ref{sec_n-gram_block}, the new method is compared with two other methods in Table~\ref{table.n-gram.ablation}: 1) the one implemented by FairSeq (called baseline). 2) a revised CPU-based kernel, which improves baseline by moving data from GPU to CPU before computing to avoid multiple data transfers (called CPU kernel). The time difference (4477.1 ms vs 584.9 ms) between baseline and CPU kernel indicates that data transfer optimization alone can speedup about 8x. Furthermore, the proposed GPU kernel, which avoids data transfer and uses parallel computation has about 75x speed gain compared to CPU kernel. 
As shown in Figure~\ref{figure.after_opt}, the computing time after optimization becomes quite small, from about 25\% to 1\% of the overall time.

\begin{table}
\small
\centering
\begin{tabular}{l|c}
\hline
Method & Time (ms) \\
\hline
baseline & 4477.1 \\
\hline
CPU kernel & 584.9 \\
\hline
GPU kernel  & 7.8 \\
\hline
\end{tabular}
\caption{\label{table.n-gram.ablation} Compare three implementations of no-repeat n-gram.} 
\end{table}

\subsection{Analysis of Asynchronous Pipeline with Parallel I/O}
Table~\ref{table.perf.analysis} measures the performances of the synchronized pipeline with single process implemented by FairSeq and the proposed asynchronous pipeline with parallel I/O in FastSeq. The throughput is increased from 2.4 samples/s to 3.6 samples/s (around 1.5x). The speedup comes from the better resource scheduling, where the asynchronous pipeline allows post-process to run in the background when running the model inference, and the support of multi-thread detokenization.
As shown in Figure~\ref{figure.after_opt}, the post-process unique time is reduced from about 38\% to 1\% of the overall time.

\subsection{Analysis of Generation Quality}
\begin{table}
\small
\centering
\setlength\tabcolsep{3pt}
\begin{tabular}{l@{\hskip3pt}ccc}
\hline 
 Model & With  &  Baseline & FastSeq  \\
   & fp16 &  R-1/R-2/R-L & R-1/R-2/R-L  \\ 
\hline 
UniLM$_{\textrm{large}}$\tablefootnote{The differences between the ROUGE scores for UniLM are due to the differences in the data preprocess and the implementations of length-penalty. } & \xmark & 43.08/20.43/40.34 & 43.09/20.29/40.32 \\
UniLM$_{\textrm{large}}$ & \cmark & 43.06/20.42/40.32 & 43.08/20.29/40.32 \\
BART$_{\textrm{large}}$ & \xmark & 44.21/21.20/41.03 & 44.21/21.20/41.03  \\
BART$_{\textrm{large}}$ & \cmark & 44.22/21.20/41.04 & 44.22/21.21/41.03  \\
ProphetNet$_{\textrm{large}}$ & \xmark & 44.20/21.17/41.30 & 44.20/21.17/41.30 \\
ProphetNet$_{\textrm{large}}$ & \cmark & 44.17/21.17/41.28 & 44.17/21.17/41.28 \\
\hline
\end{tabular}
\caption{\label{table.quality}  Metrics (ROUGE-1, ROUGE-2, and ROUGE-L) on CNN/DailyMail test set.  }
\end{table}
All optimizations in FastSeq do not affect the model generation quality. As discussed in Section~\ref{optimizations}, the logic for detecting the repeated n-gram blocks is the same for the CPU-based and GPU-based kernels, and the asynchronous pipeline with Parallel I/O only optimizes the I/O efficiency, so these two optimizations do not change the model outputs in any fashion. For the attention cache optimization, it does not affect model outputs in theory. However, in practice, if using mix precision (e.g., floating point 16) for inference, there may be a few trivial differences in the outputs due to the numerical stability issue in GPU. Similar differences can be observed when changing batch size during floating point 16 inference. But if using floating point 32, the generated results are exactly the same. That means the minor differences are not caused by the proposed cache optimization itself. In FastSeq, the unit tests have been developed to make sure the inference outputs are the same with and without FastSeq when using floating point 32. We also compare the output quality based on the CNN/DailyMail dataset (Table~\ref{table.quality}). The quite similar ROUGE scores demonstrate that FastSeq does not impact the model quality. 

\section{Related Work}
A variety of efforts have been developed to improve the efficiency of Transformer models. From the perspective of model architectures, there are efforts on reducing attention matrix size by chunking input sequences into blocks \citep{beltagy2020longformer}, or using strided convolution over the keys and queries to compress memory \citep{j.2018generating}. 
Another kind of approaches focus on reducing model size and memory consumption by weight quantization \citep{zafrir2019q8bert}, weight sharing \citep{dehghani2018universal}, and weight pruning \citep{NEURIPS2019_2c601ad9}. Knowledge distillation is another popular approach \citep{hinton2015distilling}.

On the other hand, a dozen of innovations on infrastructure side have been conducted to speed up serving of Transformer models. The fused chains of basic operators in the attention layers have been widely adopted in many frameworks (e.g., Onnx Runtime \footnote{\url{https://github.com/microsoft/onnxruntime}}, Deep Speed\footnote{\url{https://www.deepspeed.ai}}). 
It is also performance critical to optimize data layout and movement among the connected operations \citep{ivanov2020data}. In situation of varied input lengths, TurboTransformers \citep{fang2021turbotransformers} is developed to better serve online models by using dynamic batch scheduler, more efficient memory allocation and deallocation algorithms. FasterTransformers\footnote{\href{https://github.com/NVIDIA/DeepLearningExamples/tree/master/FasterTransformer}{FasterTransformer Github}} deeply optimizes kernels of encoder, decoder and beam search to better utilize computer power of Tensor Core.

\section{Conclusion}
In this work, we present FastSeq, which provides general solutions for speeding up the sequence generation without accuracy loss. 
The proposed optimizations include an attention cache optimization, an GPU-based n-grams blocking algorithm, and an asynchronous generation pipeline.
In the future, we will support more models and explore more techniques to accelerate the generation speed.


\bibliographystyle{acl_natbib}
\bibliography{acl2021}

\appendix
\section{Cache Optimization in Self-Attention}

First, we can split the cached key and value to two parts: $Cache\_K'$ and $Cache\_V'$ are for the prefix; $Cache\_K_{t}$ and $Cache\_V_{t}$ are for the generated sequence at the $t$ step as below:

\begin{equation} \tag{2}
\begin{aligned}
\underset{\scriptscriptstyle{[B, 1, N, D]}}{Cache\_K'} = \underset{\scriptscriptstyle{[B \times 1, N, D]}}{X} {W_{k}} \\
\underset{\scriptscriptstyle{[B, 1, N, D]}}{Cache\_V'} = {X} {W_{v}} \\
\underset{\scriptscriptstyle{[B \times M, t, D]}}{K_t} = concat(\underset{\scriptscriptstyle{[B \times M, t-1, D]}}{Cache\_K_{t-1}}, {y_{t-1}} \cdot \underset{\scriptscriptstyle{[D, D]}}{W_{k}}) \\
\underset{\scriptscriptstyle{[B \times M, t, D]}}{V_t} = concat(\underset{\scriptscriptstyle{[B \times M, t-1, D]}}{Cache\_V_{t-1}}, {y_{t-1}} \cdot \underset{\scriptscriptstyle{[D, D]}}{W_{v}}) \\
\end{aligned}
\label{eq.self_attn_opt.1} 
\end{equation}

The above split operation results in incompatible shapes between $Cache\_K'$ and $Cache\_K_{t}$, and between $Cache\_V'$ and $Cache\_V_{t}$. Instead of reorganizing these cached keys and values, Equation~\eqref{eq.self_attn_opt.2} is leveraged to compute $attn_t$. By this way, the expensive $concat$ operations on large tensors can be avoided.
\begin{equation} \tag{3}
\begin{aligned}
\underset{\scriptscriptstyle{[B \times M, 1, N]}}{attn\_w_0} =  einsum({Q_t}, {Cache\_K'}) \\
\underset{\scriptscriptstyle{[B \times M, 1, t]}}{attn\_w_1} = {Q_t} \cdot {K_t^T} \\
\underset{\scriptscriptstyle{[B \times M, 1, N + t]}}{attn\_w} = concat(attn\_w_0, attn\_w_1) \\
\underset{\scriptscriptstyle{[B \times M, 1, N + t]}}{attn\_prob} = softmax(\frac{attn\_w}{\sqrt{d_{k_t}}}) \\
\underset{\scriptscriptstyle{[B \times M, 1, N]}}{attn\_prob_0}, \underset{\scriptscriptstyle{[B, M, 1, t]}}{attn\_prob_1} = split({attn\_prob}) \\
\underset{\scriptscriptstyle{[B \times M, 1, D]}}{{attn_t}_0} = einsum({attn\_prob_0},{Cache\_V'}) \\
\underset{\scriptscriptstyle{[B \times M, 1, D]}}{{attn_t}_1} = {attn\_prob_1} \cdot {V_t} \\
\underset{\scriptscriptstyle{[B \times M, 1, D]}}{attn_t} = {{attn_t}_0} + {{attn_t}_1} \\
\end{aligned}
\label{eq.self_attn_opt.2} 
\end{equation}

\section{Cache Optimization in Encoder-Decoder Attention}

The first step is to remove the duplication in {Cache\_K} and {Cache\_V}. For the incompatible shape between {Q} and {Cache\_K}, $einsum$ is leveraged to avoid the reshape.
\begin{equation} \tag{4}
\begin{aligned}
\underset{\scriptscriptstyle{[B, 1, N, D]}}{Cache\_K} = \underset{\scriptscriptstyle{[B, 1, N, D]}}{S} \cdot {W_{k}} \\
\underset{\scriptscriptstyle{[B, 1, N, D]}}{Cache\_V} = {S} \cdot {W_{v}} \\
\underset{\scriptscriptstyle{[B \times M, 1, N]}}{attn\_w} = einsum({Q_{t}}, {Cache\_K}) \\
\underset{\scriptscriptstyle{[B \times M, 1, N]}}{attn\_prob_t} = softmax(\dfrac{attn\_w}{\sqrt{d_{k_{t}}}}) \\
\underset{\scriptscriptstyle{[B \times M, 1, D]}}{attn_t} = einsum({attn\_prob_t}, {Cache\_V}) \\
\end{aligned}
\label{eq.encoder_decoder_attn_opt} 
\end{equation}
As such, the size of $Cache\_K$ and $Cache\_V$ can be reduced by $M$ times from $B \times M \times N \times D$ to $B \times 1 \times N \times D$. Then the optimization benefits in self-attention can be achieved here as well, including peak memory reduction and larger batch size. Additionally, the cached key and value are not needed to be reordered since the elements at the beam dimension are exactly the same.

\end{document}